\documentclass[10pt,twocolumn,letterpaper]{article}

\usepackage[pagenumbers]{cvpr} 
\usepackage{amsmath,amssymb}

\usepackage{booktabs}
\usepackage{multirow}
\usepackage{multicol}
\usepackage[mathscr]{euscript}
\usepackage[dvipsnames]{xcolor}
\usepackage{colortbl}
\usepackage{makecell}
\usepackage{subcaption}
\usepackage[symbol]{footmisc}

\newcommand{\et}[2]{${#1}^{\pm{#2}}$}
\newcommand{\etb}[2]{$\mathbf{{#1}}^{\pm{#2}}$}
\newcommand{\ets}[2]{$\underline{{#1}}^{\pm{#2}}$}

\usepackage[dvipsnames]{xcolor}

\definecolor{cvprblue}{rgb}{0.21,0.49,0.74}
\usepackage[pagebackref,breaklinks,colorlinks,citecolor=cvprblue]{hyperref}

\def\paperID{8655} 
\def\confName{CVPR}
\def\confYear{2024}

\begin{document}
\title{MoMask: Generative Masked Modeling of 3D Human Motions }

\author{Chuan Guo$^*$
\and Yuxuan Mu$^*$
\and Muhammad Gohar Javed$^*$
\and Sen Wang
\and Li Cheng \\
University of Alberta\\
{\tt\small {\{cguo2, ymu3, javed4, lcheng5\}}@ualberta.ca}\\
{\tt\small \href{https://ericguo5513.github.io/momask/}{https://ericguo5513.github.io/momask/}}
}
\twocolumn[{
\renewcommand\twocolumn[1][]{#1}
\maketitle
\begin{center}
    \centering
    \captionsetup{type=figure}
    \includegraphics[width=0.95\linewidth]{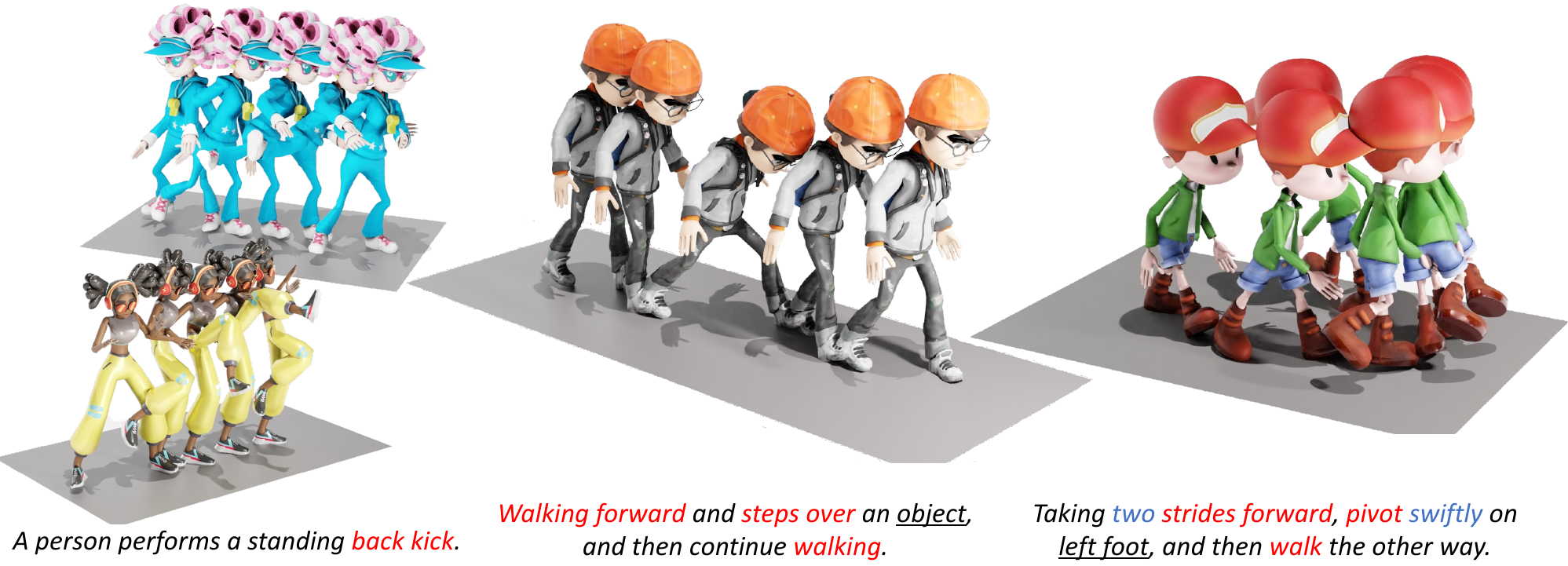}

    \captionof{figure}{Our MoMask, when provided with a text input, generates high-quality 3D human motion with diversity and precise control over subtleties such as "\textit{two strides forward}", "\textit{pivot on left foot}", and "\textit{pivot swiftly}".}
    \label{fig:teaser_image}
\end{center}
}]
\footnotetext[1]{These authors contributed equally to this work.}
\begin{abstract}
We introduce MoMask, a novel masked modeling framework for text-driven 3D human motion generation. In MoMask, a hierarchical quantization scheme is employed to represent human motion as multi-layer discrete motion tokens with high-fidelity details. Starting at the base layer, with a sequence of motion tokens obtained by vector quantization, the residual tokens of increasing orders are derived and stored at the subsequent layers of the hierarchy. This is consequently followed by two distinct bidirectional transformers. For the base-layer motion tokens, a Masked Transformer is designated to predict randomly masked motion tokens conditioned on text input at training stage. During generation (i.e. inference) stage, starting from an empty sequence, our Masked Transformer iteratively fills up the missing tokens; Subsequently, a Residual Transformer learns to progressively predict the next-layer tokens based on the results from current layer. 
Extensive experiments demonstrate that MoMask outperforms the state-of-art methods on the text-to-motion generation task, with an FID of 0.045 (vs e.g. 0.141 of T2M-GPT) on the HumanML3D dataset, and 0.228 (vs 0.514) on KIT-ML, respectively. MoMask can also be seamlessly applied in related tasks without further model fine-tuning, such as text-guided temporal inpainting. 
\end{abstract}    

\section{Introduction}
\label{sec:intro}


Generating 3D human motions from textual descriptions, aka text-to-motion generation, is a relatively new task that may play an important role in a broad range of applications such as video games, metaverse, and virtual reality \& augmented reality. In the past few years, it has generated intensive research interests~\cite{guo2022generating,petrovich2022temos,tevet2022human,chen2023executing,zhang2022motiondiffuse,zhang2023t2m,guo2022tm2t,jiang2023motiongpt,kong2023priority}. 
Among them, it has become popular to engage generative transformers in modeling human motions~\cite{guo2022tm2t,gong2023tm2d,jiang2023motiongpt,zhang2023t2m}. In this pipeline, motions are transformed into discrete tokens through vector quantization (VQ), then fed into e.g. an autoregressive model to generate the sequence of motion tokens in an unidirectional order. Though achieving impressive results, these methods shares two innate drawbacks. To begin with, the VQ process inevitably introduces approximation errors, which imposes undesired limit to the motion generation quality. Moreover, the unidirectional decoding may unnecessarily hinder the expressiveness of the generative models. For instance, consider the following scenario: at each time step, the motion content is generated by only considering the preceding (rather than global) context; furthermore, errors will often accumulate over the generation process. Though several recent efforts using discrete diffusion models~\cite{lou2023diversemotion,kong2023priority} have considered to decode the motion tokens bidirectionally, by relying on a cumbersome discrete diffusion process, they typically require hundreds of iterations to produce a motion sequence. 

Motivated by these observations, we propose a novel framework, MoMask, for high-quality and efficient text-to-motion generation by leveraging the residual vector quantization (RVQ) techniques~\cite{zeghidour2021soundstream,borsos2023audiolm,martinez2014stacked} and the recent generative masked transformers~\cite{chang2023muse,li2023mage,chang2022maskgit,yu2023magvit}. Our approach builds on the following three components. First, an RVQ-VAE is learned to establish precise mappings between 3D motions and the corresponding sequences of discrete motion tokens. Unlike previous motion VQ tokenizers~\cite{guo2022tm2t,gong2023tm2d,zhang2023t2m} that typically quantize latent codes in a single pass, our hierarchical RVQ employs iterative rounds of residual quantization to progressively reduce quantization errors. This results in multi-layer motion tokens, with the base layer serving to perform standard motion quantization, and the rest layers in the hierarchy capturing the residual coding errors of their respective orders, layer by layer. Our quantization-based hierarchical design is further facilitated by two distinct transformers, the Masked Transformer (i.e. M-Transformer) and Residual Transformer (R-Transformer), that are dedicated to generating motion tokens for the base VQ layer and the rest residual layers, respectively.

The M-Transformer, based on BERT~\cite{devlin2018bert}, is trained to predict the randomly masked tokens at the base layer, conditioned on textual input. The ratio of masking, instead of being fixed~\cite{devlin2018bert, he2022masked}, is a scheduled variable that ranges from 0 to 1. During generation, starting from all tokens being masked out, M-Transformer produces a complete sequence of motion tokens within a small number of iterations. At each iteration, all masked tokens are predicted \textit{simultaneously}. Predicted tokens with the highest confidence will remain unchanged, while the others are masked again and re-predicted in the next iteration. Once the base-layer tokens are generated, the R-Transformer ensues to progressively predict the residual tokens of the subsequent layer given the token sequence at current layer. Overall, the entire set of layered motion tokens can be efficiently generated within merely 15 iterations, regardless of the motion's length.


Our main contributions can be summarized as follows: First, our MoMask is the first generative masked modeling framework for the problem of text-to-motion generation. It comprises of a hierarchical quantization generative model and the dedicated mechanism for precise residual quantization, base token generation and residual token prediction. Second, our MoMask pipeline produces precise and efficient text-to-motion generation. Empirically, it achieves new state-of-the-art performance on text-to-motion generation task with an FID of 0.045 (\textit{vs.} 0.141 in~\cite{zhang2023t2m}) on HumanML3D and 0.204 (\textit{vs.} 0.514 in~\cite{zhang2023t2m}) on KIT-ML. Third, our MoMask also works well for related tasks, such as text-guided motion inpainting.

\begin{figure*}[h]
	\centering
	\includegraphics[width=\linewidth]{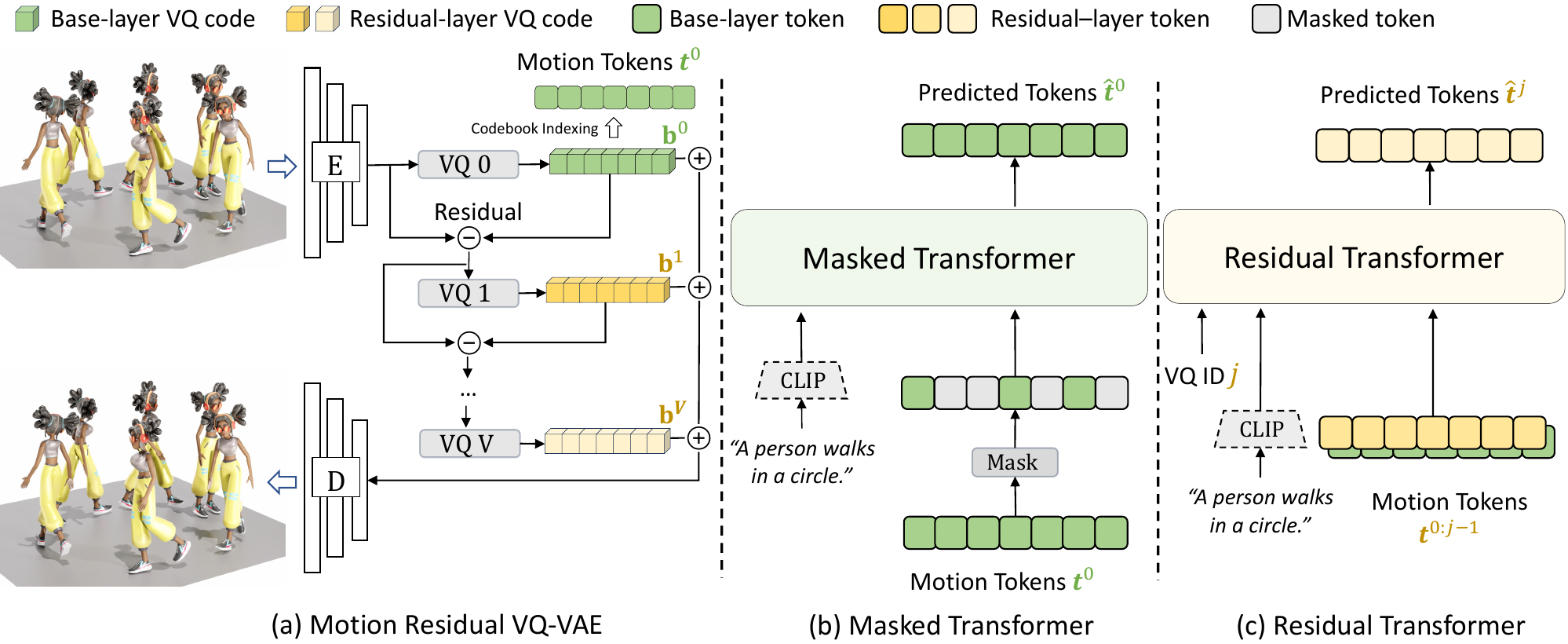}
	\caption{\textbf{Approach overview.} (a) Motion sequence is tokenized through vector quantization (VQ), also referred to as the base quantization layer, as well as a hierarchy of multiple layers for residual quantization. (b) \textit{Parallel} prediction by the Masked Transformer: the tokens in the base layer \textcolor{LimeGreen}{$t^0$} are randomly masked out with a variable rate, and then a text-conditioned masked transformer is trained to predict the masked tokens in the sequence simultaneously. (c) Layer-by-layer \textit{progressive} prediction by the Residual Transformer. A text-conditioned residual transformer learns to progressively predict the residual tokens \textcolor{Tan}{$t^{j>0}$} from the tokens in previous layers, $t^{0:j-1}$. }
	\label{fig:model}
\end{figure*}

\section{Related Work}
\label{sec:related_work}

\noindent\textbf{Human Motion Generation}. Recently, we have witnessed the surge of works for neural motion generation, with conditioning on various domains such as motion prefix~\cite{mao2019learning,liu2022investigating}, action class~\cite{guo2020action2motion,petrovich2021action,cervantes2022implicit,lucas2022posegpt}, audio~\cite{gong2023tm2d,zhou2023ude,siyao2022bailando,tseng2023edge}, texts~\cite{guo2022tm2t,petrovich2022temos,guo2022generating,tevet2022human,chen2023executing}. Early works~\cite{ahuja2019language2pose,ghosh2021synthesis,plappert2018learning,lin2018generating,huang2020dance} commonly model motion generation deterministically, resulting in averaged and blurry motion results. This is properly addressed by stochastic models. GAN modeling is adopted in~\cite{cai2018deep,wang2020learning} for action-conditioned motion generation. Meanwhile, temporal VAE framework and transformer architecture are exploited in the works of  ~\cite{guo2022action2video,petrovich2021action}. In terms of text-to-motion generation, T2M~\cite{guo2022generating} extended the temporal VAE to learn the probabilistic mapping between texts and motions. Similarly, TEMOS~\cite{petrovich2022temos} takes advantage of Transformer VAE to optimize a joint variational space between natural language and motions, which is extended by TEACH~\cite{athanasiou2022teach} for long motion compositions. MotionCLIP~\cite{tevet2022motionclip} and ohMG~\cite{lin2023being} model text-to-motion in an unsupervised manner using the large pretrained CLIP~\cite{radford2021learning} model. The emerging diffusion models and autoregressive models have significantly changed the field of motion generation. In diffusion methods, a network is learned to gradually denoise the motion sequence, supervised by a scheduled diffusion process~\cite{tevet2022human,kim2023flame,zhang2022motiondiffuse,tseng2023edge,chen2023executing,kong2023priority,lou2023diversemotion}. 
Regarding autoregressive models~\cite{guo2022tm2t,zhang2023t2m,jiang2023motiongpt,zhang2023motiongpt,gong2023tm2d}, motions are firstly discretized as tokens via vector quantization~\cite{van2017neural}, which are then modeled by the expressive transformers as in language model.


\noindent\textbf{Generative Masked Modeling.}
BERT~\cite{devlin2018bert} introduces masked modeling for language tasks that word tokens are randomly masked out with a fixed ratio, and then the bi-directional transformer learns to predict the masked tokens. Despite being a decent pre-trained text encoder, BERT cannot synthesize novel samples. In this regard, ~\cite{chang2022maskgit} proposes to mask the tokens with a variable and traceable rate that is controlled by a scheduling function. Therefore, new samples can be synthesized iteratively following the scheduled masking. MAGE~\cite{li2023mage} unifies representation learning and image synthesis using the masked generative encoder. Muse~\cite{chang2023muse} extends this paradigm for text-to-image generation and editing. Magvit~\cite{yu2023magvit} suggests a versatile masking strategy for multi-task video generation. Inspired by these successes, we first introduce generative masked modeling for human motion synthesis in this paper.

\noindent\textbf{Deep Motion Quantization and RVQ.}
~\cite{aristidou2018deep} learns semantically meaningful discrete motif words leveraging triplet contrastive learning. TM2T~\cite{guo2022tm2t} starts applying vector quantized-VAE~\cite{van2017neural} to learn the mutual mapping between human motions and discrete tokens, where the autoencoding latent codes are replaced with the selected entries from a codebook. T2M-GPT~\cite{zhang2023t2m} further enhances the performance using EMA and code reset techniques. Nevertheless, the quantization process inevitably introduces errors, leading to suboptimal motion reconstruction.
In this work, we adapt residual quantization~\cite{zeghidour2021soundstream,borsos2023audiolm,martinez2014stacked}, a technique used in neural network compression~\cite{li2021trq,li2017performance,ferdowsi2017regularized} and audio quantization~\cite{borsos2023audiolm, wang2023neural} which iteratively quantizes a vector and its residuals. This approach represents the vector as a stack of codes, enabling high-precision motion discretization.

\begin{figure*}[th]
	\centering
	\includegraphics[width=\linewidth]{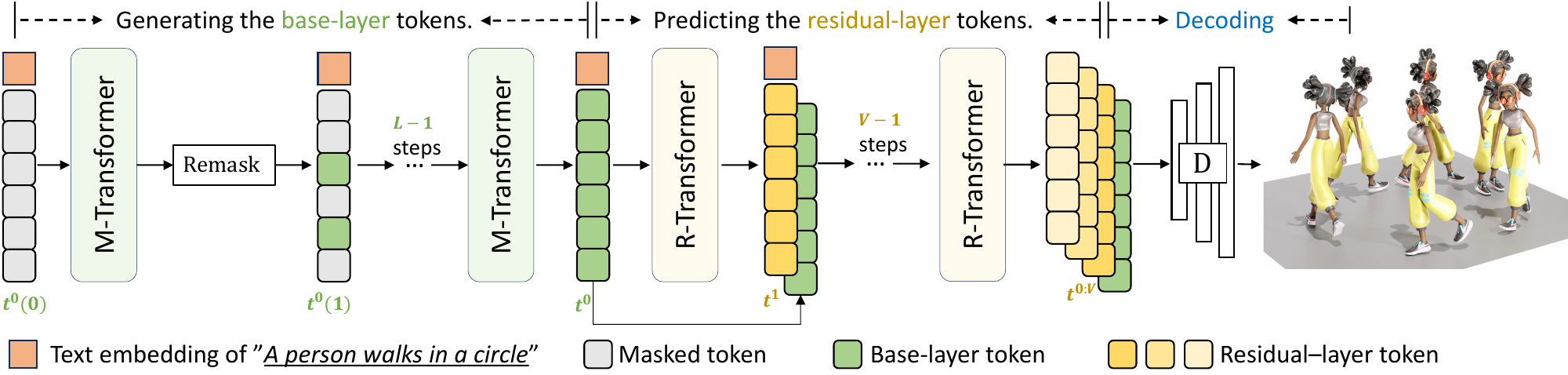}
	\caption{\textbf{Inference process.} Starting from an empty sequence $t^0(0)$, the M-Transformer generates the base-layer token sequence $t^0$ in \textcolor{Tan}{$L$} iterations. Following this, the R-Transformer progressively predicts the rest-layer token sequences $t^{2:V}$ within \textcolor{Dandelion}{$V-1$} steps.}
	\label{fig:inference}
\end{figure*}
\section{Approach}
\label{sec:approach}
Our goal is to generate a 3D human pose sequence $\mathbf{m}_{1:N}$ of length $N$ guided by a textual description $c$, where $\mathbf{m}_i\in\mathbb{R}^D$ with $D$ denoting the dimension of pose features. As illustrated in ~\cref{fig:model}, our approach consists of three principle components: a residual-based quantizer that tokenizes motion sequence into multi-layer discrete tokens (\cref{subsec:RVQ}), a masked transformer that generates motion tokens in the base layer (\cref{subsec:masktransformer}), and a residual transformer (\cref{subsec:restransformer}) that predicts the tokens in the subsequent residual layers. The inference process of generation is detailed in~\cref{subsec:inference}.

\subsection{Training: Motion Residual VQ-VAE}
\label{subsec:RVQ}
Conventional motion VQ-VAEs~\cite{guo2022tm2t,zhang2023t2m,jiang2023motiongpt,zhang2023motiongpt} transform a motion sequence into one tuple of discrete motion tokens. Specifically, the motion sequence $\mathbf{m}_{1:N}\in\mathbb{R}^{N\times  D}$ is firstly encoded into a latent vector sequence $\mathbf{\Tilde{b}}_{1:n}\in\mathbb{R}^{n\times d}$ with downsampling ratio of $n/N$ and latent dimension $d$, using 1D convolutional encoder $\mathrm{E}$; 
each vector is subsequently replaced with its nearest code entry in a preset codebook $\mathcal{C}=\{\mathbf{c}_k\}_{k=1}^{K}\subset \mathbb{R}^d$, known as quantization $\mathrm{Q}(\cdot)$. Then the quantized code sequence $\mathbf{b}_{1:n}=\mathrm{Q}(\mathbf{\Tilde{b}}_{1:n})\in\mathbb{R}^{n\times d}$ is projected back to motion space for reconstructing the motion $\mathbf{\hat{m}=\mathrm{D}(\mathbf{b})}$. After all, the indices of the selected codebook entries (namely \textit{motion tokens}) are used as the alternative discrete representation of input motion. Though effective, the quantization operation $\mathrm{Q}(\cdot)$ inevitably leads to information loss, which further limits the quality of reconstruction. 

To address this issue, we introduce residual quantization (RQ) as described in \cref{fig:model}(a). 
In particular, RQ represents a motion latent sequence $\mathbf{\Tilde{b}}$ as $V+1$ \textit{ordered} code sequences, using $V+1$ quantization layers. Formally, this is defined as $\mathrm{RQ}(\mathbf{\Tilde{b}}_{1:n})=\left[\mathbf{b}_{1:n}^v\right]_{v=0}^V$,
with $\mathbf{b}_{1:n}^v \in \mathbb{R}^{n\times d}$ denoting the code sequence at the $v$-th quantization layer. Concretely, starting from $0$-th residual $\mathbf{r}^0=\mathbf{\Tilde{b}}$, RQ recursively calculates $\mathbf{b}^v$ as the approximation of residual $\mathbf{r}^{v}$, and then the next residual $\mathbf{r}^{v+1}$ as
\begin{align}
    \mathbf{b}^v = \mathrm{Q}(\mathbf{r}^{v}), \quad \mathbf{r}^{v+1} = \mathbf{r}^{v} - \mathbf{b}^{v},
\end{align}
for $v=0,...,V$. After RQ, the final approximation of latent sequence $\mathbf{\Tilde{b}}$ is the sum of all quantized sequences $\sum_{v=0}^V \mathbf{b}^v$, which is then fed into decoder $\mathrm{D}$ for motion reconstruction. 

Overall, the residual VQ-VAE is trained via a motion reconstruction loss combined with a latent embedding loss at each quantization layer:
\begin{align}
    \mathcal{L}_{rvq} = \|\mathbf{m}-\mathbf{\hat{m}}\|_1 + \beta\sum_{v=1}^V\|\mathbf{r}^{v}-\mathrm{sg}[\mathbf{b}^v]\|_2^2,
\end{align}
where $\mathrm{sg}[\cdot]$ denotes the stop-gradient operation, and $\beta$ a weighting factor for embedding constraint. This framework is optimized with straight-though gradient estimator~\cite{van2017neural}, and our codebooks are updated via exponential moving average and codebook reset following T2M-GPT~\cite{zhang2023t2m}. 

\noindent\textbf{Quantization Dropout.} Ideally, the early quantization layers are expected to restore the input motion as much as possible; then the later layers add up the missing finer details. 
To exploit the capacity of each quantizer, we adopt a \textit{quantization dropout} strategy, which randomly disables the last 0 to $V$ layers with probability $q\in [0, 1]$ during training.

After training, each motion sequence $\mathbf{m}$ can be represented as $V+1$ discrete motion token sequences $T = [t_{1:n}^v]_{v=0}^{V}$ where each token sequence $t_{1:n}^v \in \{1,...,|\mathcal{C}^v|\}^n$ is the ordered codebook-indices of quantized embedding vectors $\mathbf{b}_{1:n}^v$, such that $\mathbf{b}_i^v = \mathcal{C}_{t_i^v}^v$ for $i\in[1, n]$. Among these $V+1$ sequences, the first (i.e. base) sequence possesses the most prominent information, while the impact of subsequent layers gradually diminishes.

\subsection{Training: Masked Transformer}
\label{subsec:masktransformer}
Our bidirectional masked transformer is designed to model the base-layer motion tokens $t_{1:n}^0\in\mathbb{R}^n$, as illustrated in~\Cref{fig:model}(b). We first randomly masked out a varying fraction of sequence elements, by replacing the tokens with a special $[\mathrm{MASK}]$ token. With $\Tilde{t}^0$ denoting the sequence after masking, the goal is to predict the masked tokens given text $c$ and $\Tilde{t}^0$. We use CLIP~\cite{radford2021learning} for extracting text features. Mathematically, our masked transformer $p_\theta$ is optimized to minimize the negative log-likelihood of target predictions:

\begin{align}
    \mathcal{L}_{mask} = \sum_{\Tilde{t}_k^0=[\mathrm{MASK}]}-\log p_\theta(t_k^0| \Tilde{t}^0, c).
\end{align}

\noindent\textbf{Mask Ratio Schedule.} We adopt a cosine function $\gamma(\cdot)$ for scheduling the masking ratio following~\cite{chang2023muse,chang2022maskgit}. Practically, the mask ratio is obtained by $\gamma(\tau)=\cos(\frac{\pi\tau}{2})\in[0, 1]$, where $\tau\in[0, 1]$ that $\tau=0$ means the sequence is completely corrupted. 
During training, the $\tau\sim\mathcal{U}(0, 1)$ is randomly sampled, and then $m=\lceil\gamma(\tau)\cdot n\rceil$ sequence entries are uniformly selected to be masked with $n$ denoting the length of sequence.

\noindent\textbf{Replacing and Remasking.} To enhance the contextual reasoning of the masked transformer, we adopt the remasking strategy used in BERT pretraining~\cite{devlin2018bert}. If a token is selected for masking, we replace this token with (1) $[\mathrm{MASK}]$ token 80\% of the time; (2) a random token 10\% of the time; and (3) an unchanged token 10\% of the time.

\begin{table*}[thb]
    \centering
    \scalebox{0.9}{
    \begin{tabular}{l l c c c c c c}
    \toprule
    \multirow{2}{*}{Datasets} & \multirow{2}{*}{Methods}  & \multicolumn{3}{c}{R Precision$\uparrow$} & \multirow{2}{*}{FID$\downarrow$} & \multirow{2}{*}{MultiModal Dist$\downarrow$} & \multirow{2}{*}{MultiModality$\uparrow$}\\

    \cline{3-5}
       ~& ~ & Top 1 & Top 2 & Top 3 \\
    \midrule
    \multirow{9}{*}{\makecell[c]{Human\\ML3D}} & TM2T~\cite{guo2022tm2t} & \et{0.424}{.003} & \et{0.618}{.003} & \et{0.729}{.002} & \et{1.501}{.017} & \et{3.467}{.011} & \ets{2.424}{.093}  \\ 
        ~& T2M~\cite{guo2022generating} & \et{0.455}{.003} & \et{0.636}{.003} & \et{0.736}{.002} & \et{1.087}{.021} & \et{3.347}{.008} & \et{2.219}{.074}  \\   
        ~& MDM~\cite{tevet2022human} & - & - & \et{0.611}{.007} & \et{0.544}{.044} & \et{5.566}{.027} & \etb{2.799}{.072}  \\

        ~ & MLD~\cite{chen2023executing} & \et{0.481}{.003} & \et{0.673}{.003} & \et{0.772}{.002} & \et{0.473}{.013} & \et{3.196}{.010} & \et{2.413}{.079}  \\
        ~ & MotionDiffuse~\cite{zhang2022motiondiffuse} & \et{0.491}{.001} & \et{0.681}{.001} & \et{0.782}{.001} & \et{0.630}{.001} & \et{3.113}{.001}  & \et{1.553}{.042}  \\

        ~ & T2M-GPT~\cite{zhang2023t2m} & \et{0.492}{.003} & \et{0.679}{.002} & \et{0.775}{.002} & \et{0.141}{.005} & \et{3.121}{.009}  & \et{1.831}{.048}  \\

        ~ & ReMoDiffuse~\cite{zhang2023remodiffuse} & \ets{0.510}{.005} & \et{0.698}{.006} & \et{0.795}{.004} & \et{0.103}{.004} & \ets{2.974}{.016} & \et{1.795}{.043}  \\
    \cline{2-8}
        ~ & MoMask (base) & \et{0.504}{.004} & \ets{0.699}{.006} & \ets{0.797}{.004} & \ets{0.082}{.008} & \et{3.050}{.013}  & \et{1.050}{.061}  \\
        ~ & \textbf{MoMask} & \etb{0.521}{.002} & \etb{0.713}{.002} & \etb{0.807}{.002} & \etb{0.045}{.002} & \etb{2.958}{.008} & \et{1.241}{.040}  \\
    \midrule
    \multirow{9}{*}{\makecell[c]{KIT-\\ML}} & TM2T~\cite{guo2022tm2t} & \et{0.280}{.005} & \et{0.463}{.006} & \et{0.587}{.005} & \et{3.599}{.153} & \et{4.591}{.026} & \etb{3.292}{.081}  \\ 
        ~& T2M~\cite{guo2022generating} & \et{0.361}{.005} & \et{0.559}{.007} & \et{0.681}{.007} & \et{3.022}{.107} & \et{3.488}{028} & \et{2.052}{.107}  \\   
        ~& MDM~\cite{tevet2022human} & - & - & \et{0.396}{.004} & \et{0.497}{.021} & \et{9.191}{.022} & \et{1.907}{.214}  \\

        ~& MLD~\cite{chen2023executing} & \et{0.390}{.008} & \et{0.609}{.008} & \et{0.734}{.007} & \et{0.404}{.027} & \et{3.204}{.027} & \ets{2.192}{.071}  \\
        
        ~& MotionDiffuse~\cite{zhang2022motiondiffuse} & \et{0.417}{.004} & \et{0.621}{.004} & \et{0.739}{.004} & \et{1.954}{.062} & \et{2.958}{.005}  & \et{0.730}{.013}  \\

        ~& T2M-GPT~\cite{zhang2023t2m} & \et{0.416}{.006} & \et{0.627}{.006} & \et{0.745}{.006} & \et{0.514}{.029} & \et{3.007}{.023}  & \et{1.570}{.039}  \\

        ~& ReMoDiffuse~\cite{zhang2023remodiffuse} & \ets{0.427}{.014} & \ets{0.641}{.004} & \ets{0.765}{.055} & \etb{0.155}{.006} & \ets{2.814}{.012} & \et{1.239}{.028}  \\
    \cline{2-8}
        ~& MoMask (base) & \et{0.415}{.010} & \et{0.634}{.011} & \et{0.760}{.005} & \et{0.372}{.020} & \et{2.931}{.041}  & \et{1.097}{.054}  \\
        ~& \textbf{MoMask} & \etb{0.433}{.007} & \etb{0.656}{.005} & \etb{0.781}{.005} & \ets{0.204}{.011} & \etb{2.779}{.022} & \et{1.131}{.043}  \\
    \bottomrule
    \end{tabular}
    }
    \caption{\textbf{Quantitative evaluation on the HumanML3D and KIT-ML test set.} $\pm$ indicates a 95\% confidence interval. MoMask (base) means that MoMask only uses base-layer tokens. \textbf{Bold} face indicates the best result, while \underline{underscore} refers to the second best.}
    \label{tab:quantitative_eval}

\end{table*}
\subsection{Training: Residual Transformer}
\label{subsec:restransformer}
We learn a single residual transformer to model the tokens from the other $V$ residual quantization layers. The residual transformer has a similar architecture to the masked transformer (\cref{subsec:masktransformer}), except that it contains $V$ separate embedding layers. During training, we randomly select a quantizer layer $j\in[1, V]$ to learn. All the tokens in the preceding layers $t^{0:j-1}$ are embedded and summed up as the token embedding input. Taking the token embedding, text embedding, and RQ layer indicator $j$ as input, the residual transformer $p_\phi$ is trained to predict the $j$-th layer tokens in parallel. Overall, the training objective is:
\begin{align}
    \mathcal{L}_{res} = \sum_{j=1}^V\sum_{i=1}^n-\log p_\phi(t_i^j| t_i^{1:j-1}, c, j).
\end{align}

We also share the parameters of the $j$-th prediction layer and the $(j+1)$-th motion token embedding layer for more efficient learning.

\subsection{Inference}
\label{subsec:inference}
As presented in~\Cref{fig:inference}, there are three stages in inference. Firstly, starting from an empty sequence $t^0(0)$ that all tokens are masked out, we expect to generate the base-layer token sequence $t^0$ of length $n$ in $L$ iterations. Given the masked token sequence at $l$-th iteration $t^0(l)$, M-Transformer first predicts the probability distribution of tokens at the masked locations, and samples motion tokens with the probability. Then the sampled tokens with the lowest $\lceil\gamma(\frac{l}{L})\cdot n\rceil$ confidences are masked again, and the other tokens will remain unchanged for the rest iterations. This new token sequence $t^0(l+1)$ is used to predict the token sequence at the next iteration until $l$ reaches $L$. Once the base-layer tokens are completely generated, the R-Transformer progressively predicts the token sequence in the rest quantization layers. Finally, all tokens are decoded and projected back to motion sequences through the RVQ-VAE decoder.  

\noindent \textbf{Classifier Free Guidance.} We adopt classifier-free guidance (CFG)~\cite{chang2023muse,ho2022classifier} for the prediction of both M-Transformer and R-Transformer. During training, we train the transformers unconditionally $c=\emptyset$ with probability of $10\%$. During inference, CFG takes place at the final linear projection layer before softmax, where the final logits $\omega_g$ are computed by moving the conditional logits $\omega_c$ away from the unconditional logits $\omega_u$ with guidance scale $s$:
\begin{align}
    \omega_g = (1+s)\cdot\omega_c - s\cdot\omega_u.
\end{align}





\section{Experiments}
\label{sec:experiment}

\begin{figure*}[thb]
	\centering
	\includegraphics[width=\linewidth]{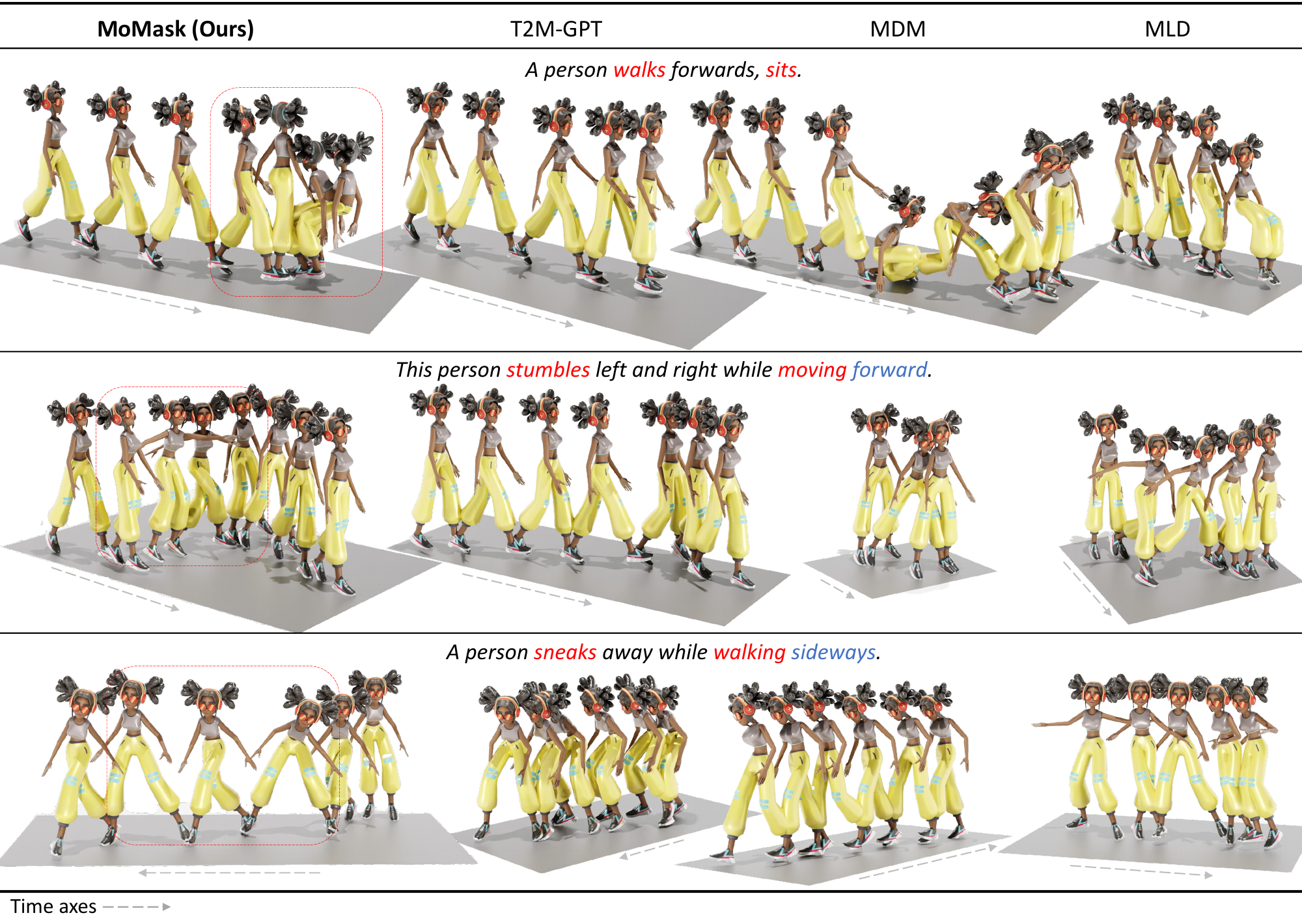}
	\caption{\textbf{Visual comparisons} between the different methods given three distinct text descriptions from HumanML3D testset. Only key frames are displayed. Compared to previous methods, MoMask generates motions with higher quality and better understanding of the subtle language concepts such as "\textit{stumble}", "\textit{sneak}", "\textit{walk sideways}". Please refer to the demo video for complete motion clips.}
        \label{fig:visual_comparison}
\end{figure*}

Empirical evaluations are conducted on two widely used motion-language benchmarks, HumanML3D~\cite{guo2022generating} and KIT-ML~\cite{plappert2016kit}. \textbf{HumanML3D} dataset collects 14,616 motions from AMASS~\cite{mahmood2019amass} and HumanAct12~\cite{guo2020action2motion} datasets, with each motion described by 3 textual scripts, totaling 44,970 descriptions. This diverse motion-language dataset contains a variety of actions, including exercising, dancing, and acrobatics. \textbf{KIT-ML} dataset consists of 3,911 motions and 6,278 text descriptions, offering an small-scale evaluation benchmark. For both motion datasets, we adopt the pose representation from the work of T2M~\cite{guo2022generating}. The datasets are augmented by mirroring, and divided into training, testing, and validation sets with the ratio of \textbf{0.8:0.15:0.05}.

\noindent\textbf{Evaluation metrics} from T2M~\cite{guo2022generating} are also adopted throughout our experiments including: (1) \textit{Frechet Inception Distance} (FID), which evaluates the overall motion quality by measuring the distributional difference between the high-level features of the generated motions and those of real motions; (2) \textit{R-Precision} and \textit{multimodal distance}, which gauge the semantic alignment between input text and generated motions; and (3) \textit{Multimodality} for assessing the diversity of motions generated from the same text. 

Though \textit{multimodality} is indeed important, we stress its role as a secondary metric that should be assessed in the conjunction with primary performance metrics such as FID and RPrecision. Emphasizing multimodality without considering the overall quality of generated results could lead to optimization of models that produce random outputs for any given input.

\noindent\textbf{Implementation Details.} Our models are implemented using PyTorch. For the motion residual VQ-VAE, we employ resblocks for both the encoder and decoder, with a downscale factor of 4. The RVQ consists of 6 quantization layers, where each layer's codebook contains 512 512-dimensional codes. The quantization dropout ratio $q$ is set to 0.2. Both the masked transformer and residual transformer are composed of 6 transformer layers, with 6 heads and a latent dimension of 384, applied to the HumanML3D and KIT-ML datasets. The learning rate reaches 2e-4 after 2000 iterations with a linear warm-up schedule for the training of all models. The mini-batch size is uniformly set to 512 for training RVQ-VAE and 64, 32 for training transformers on HumanML3D and KIT-ML, respectively. During inference, we use the CFG scale of 4 and 5 for M-Transformer and R-Transformer on HumanML3D, and (2, 5) on KIT-ML. Meanwhile, $L$ is set to 10 on both datasets.


\subsection{Comparison to state-of-the-art approaches}

We compare our approach to a set of existing state-of-the-art works ranging from VAE~\cite{guo2022generating}, diffusion-based models~\cite{tevet2022human,chen2023executing,zhang2023remodiffuse}, to autoregressive models~\cite{guo2022tm2t,zhang2023t2m}.

\begin{figure*}[thb]
	\centering
	\includegraphics[width=0.7\linewidth]{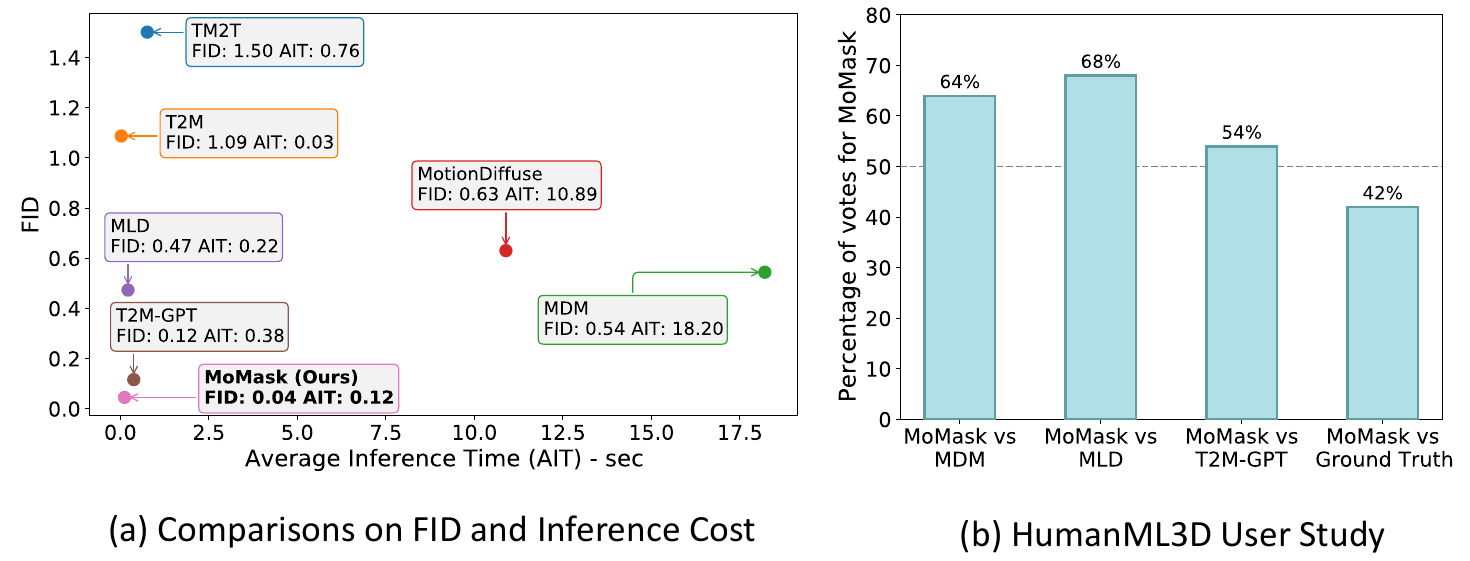}
	\caption{\textbf{(a) Comparison of inference time costs.} All tests are conducted on the same Nvidia2080Ti. The closer the model is to the origin, the better. \textbf{(b) User study results on the HumanML3D dataset.} Each bar represents the preference rate of MoMask over the compared model. Overall, MoMask is preferred over the other models most of the time. The dashed line marks 50\%.}
        \label{fig:analysis}
\end{figure*}

\noindent\textbf{Quantitative Comparisons.}
Following previous practices~\cite{guo2022generating,tevet2022human}, each experiment is repeated 20 times, and the reported metric values represent the mean with a 95\% statistical confidence interval. Additionally, we conduct experiments with MoMask exclusively generating the base-layer motion tokens, denoted as MoMask (base). Quantitative results for the HumanML3D and KIT-ML datasets are presented in \Cref{tab:quantitative_eval}.

Overall, MoMask attains state-of-the-art performance on both datasets, demonstrating substantial improvements in metrics such as FID, R-Precision, and multimodal distance. For the suboptimal performance on KIT-ML dataset, we would like to point out that the leading model, ReMoDiffuse~\cite{zhang2023remodiffuse}, involves more intricate data retrieval from a large database to achieve high-quality motion generation. Additionally, we observe that MoMask, even with the base-layer tokens alone, already achieves competitive performance compared to baselines, and the inclusion of residual tokens further elevates the results to a higher level.

In \Cref{fig:analysis}(a), we evaluate the efficiency and quality of motion generation using various methods. The inference cost is calculated as the average inference time over 100 samples on one Nvidia2080Ti device. Comparing to baseline methods, MoMask positions itself more favorably between generation quality and efficiency.

\noindent\textbf{User Study.} We further conduct a user study on Amazon Mechanical Turk to validate our previous conclusions. This user study involves 42 AMT users with \textit{master} recognition, with the side-by-side comparisons between MoMask and each of the state-of-the-art methods including MDM~\cite{tevet2022human}, MLD~\cite{chen2023executing} and T2M-GPT~\cite{zhang2023t2m}. We generate the 50 motions for each method using the same text pool from HumanML3D test set, and collect feedback from 3 distinct users for each comparison. As shown in~\cref{fig:analysis}(b), MoMask is preferred by users in most of the time, and even earns 42\% of preference on par with ground truth motions. 

\noindent\textbf{Qualitative Comparisons.} \Cref{fig:visual_comparison} displays qualitative comparisons of our approach and MDM\cite{tevet2022human}, MLD~\cite{chen2023executing}, and T2M-GPT~\cite{zhang2023t2m}. MDM~\cite{tevet2022human} usually generates overall semantically correct motions but fails to capture nuanced concepts such as "\textit{sneak}" and "\textit{sideways}". Though T2M-GPT~\cite{zhang2023t2m} and MLD~\cite{chen2023executing} have improved performance in this aspect, they still find it difficult to generate motions accurately aligned with the textual description. For example, in the bottom row, the motions from these two methods either forget to \textit{walk sideways} (T2M-GPT~\cite{zhang2023t2m}) or to \textit{sneak away} (MLD~\cite{chen2023executing}). Moreover, MLD~\cite{chen2023executing} sometimes produces lifeless motions where the character slides around, as shown in the top row. In comparison, our method is able to generate high-quality motions faithful to the input texts. Please refer to supplementary videos for dynamic visualizations.

\begin{table}[t]
    \centering
    \scalebox{0.72}{

    \begin{tabular}{l c c c c c}
    \toprule
    \multirow{2}{*}{Methods}  & \multicolumn{2}{c}{Reconstruction} & & \multicolumn{2}{c}{Generation}\\

    \cline{2-3}
    \cline{5-6}
       ~ & FID$\downarrow$ & MPJPE$\downarrow$&  & FID$\downarrow$ & MM-Dist$\downarrow$  \\
    \hline
    \rowcolor{lightgray}
    \multicolumn{6}{c}{\textit{\textbf{Evaluation on KIT-ML dataset}}} \\
    \hline
        M2DM~\cite{kong2023priority} & \et{0.413}{.009} & -& & \et{0.515}{.029} &\et{3.015}{.017} \\
        T2M-GPT~\cite{zhang2023t2m} & \et{0.472}{.011} &-& & \et{0.514}{.029} & \et{3.007}{.023} \\
        \textbf{MoMask} & \etb{0.112}{.002} & \textbf{37.2} &  & \etb{0.228}{.011} & \etb{2.774}{.022} \\
    \hline
    \rowcolor{lightgray}
    \multicolumn{6}{c}{\textit{\textbf{Evaluation on HumanML3D dataset}}} \\
    \hline
        TM2T~\cite{guo2022tm2t} &\et{0.307}{.002}& 230.1 && \et{1.501}{.017} & \et{3.467}{.011} \\
        M2DM~\cite{kong2023priority} & \et{0.063}{.001} & - && \et{0.352}{.005} &\et{3.116}{.008} \\
        T2M-GPT~\cite{zhang2023t2m} & \et{0.070}{.001} & 58.0 && \et{0.141}{.005} & \et{3.121}{.009} \\
        \textbf{MoMask} & \etb{0.019}{.001} & \textbf{29.5} && \etb{0.051}{.002} & \etb{2.957}{.008} \\
    \midrule
        \,\, \textit{w/o} RQ & \et{0.091}{.001} & 58.7 && \et{0.093}{.004} & \et{3.031}{.009} \\
        \,\, \textit{w/o} QDropout & \et{0.077}{.000} & 39.3 & & \et{0.091}{.003} & \et{2.959}{.008} \\
        \,\, \textit{w/o} RRemask & - & - & &\et{0.063}{.003} & \et{3.049}{.006} \\
    \hline
    \midrule
    \textcolor{black}{MoMask ($V$, 0)}& \et{0.091}{.001} & 58.7 &  & \et{0.093}{.004} & \et{3.031}{.009} \\
    MoMask ($V$, 1)& \et{0.069}{.001} &54.6 & &\et{0.073}{.003} & \et{3.031}{.008} \\
    MoMask ($V$, 2)& \et{0.049}{.002}  & 46.0 & & \et{0.072}{.003} & \et{2.978}{.006} \\
    MoMask ($V$, 3)& \et{0.037}{.001}  & 42.5 & & \et{0.064}{.003} & \et{2.970}{.007} \\
    MoMask ($V$, 4)& \et{0.027}{.001}  & 35.3 & & \et{0.069}{.003} & \et{2.987}{.007} \\
    MoMask ($V$, 5) & \et{0.019}{.001}  & 29.5 & & \etb{0.051}{.002} & \etb{2.962}{.008} \\
    MoMask ($V$, 6) & \et{0.014}{.001}  & 26.7 & & \et{0.076}{.003} & \et{2.994}{.007} \\
    MoMask ($V$, 7) & \etb{0.014}{.000}  & \textbf{25.3} & & \et{0.084}{.004} & \et{2.968}{.007} \\

    \midrule
    \textcolor{black}{MoMask ($q$, 0)}& \et{0.077}{.000} &  39.3 & & \et{0.091}{.003} & \et{2.959}{.008} \\
    MoMask ($q$, 0.2)& \etb{0.019}{.001} & \textbf{29.5} & & \etb{0.051}{.002} & \etb{2.957}{.008} \\
    MoMask ($q$, 0.4)& \et{0.021}{.000} &  30.2 & & \et{0.082}{.003} & \et{3.006}{.007} \\
    MoMask ($q$, 0.6)& \et{0.024}{.000} & 33.2 & & \et{0.053}{.003} & \et{2.946}{.006} \\
    MoMask ($q$, 0.8)& \et{0.023}{.000} & 33.4 & & \et{0.083}{.004} & \et{3.002}{.008} \\

    \bottomrule
    \end{tabular}
    }
    \caption{Comparison of our RVQ design vs. motion VQs from previous works~\cite{kong2023priority, zhang2023t2m,guo2022tm2t}, and further analysis on residual quantization (RQ), quantization dropout (QDropout), and replacing \& remasking (RRmask). $V$ and $q$ are the number of RQ and QDropout ratio, respectively. MPJPE is measured in millimeters.}
    \label{tab:ablation}

\end{table}

\begin{figure*}[thb]
	\centering
	\includegraphics[width=\linewidth]{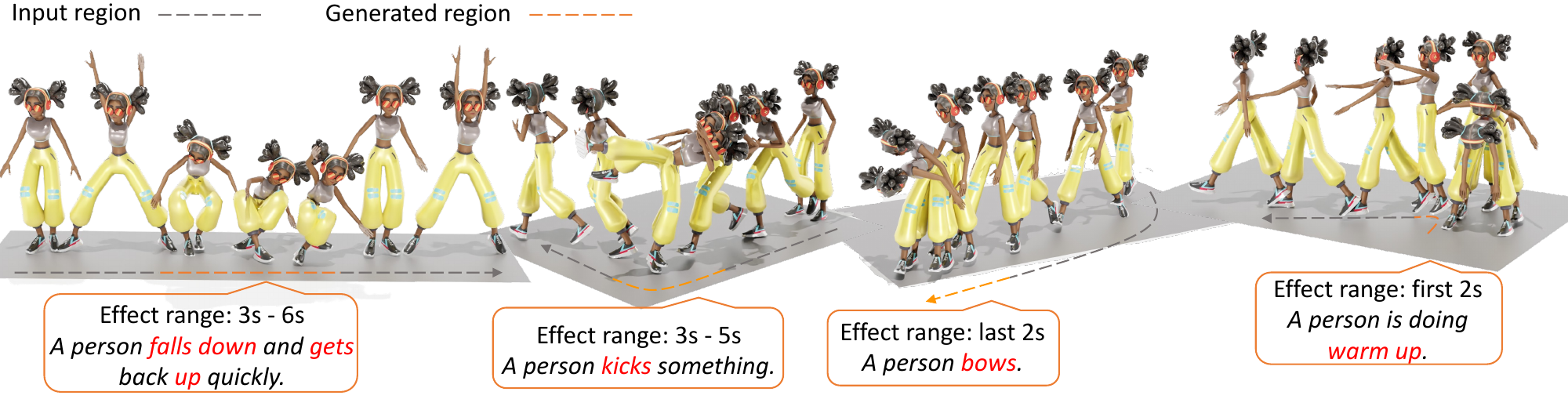}

	\caption{\textbf{Examples of temporal inpainting.} Dark dash line indicates the range(s) where the motion content(s) is given by the reference sequence. Orange dash line indicates the range of motion content generated by MoMask, conditioned on the text prompt below.}
        \label{fig:editing}
\end{figure*}

\begin{figure}[thb]
	\centering
	\includegraphics[width=0.7\linewidth]{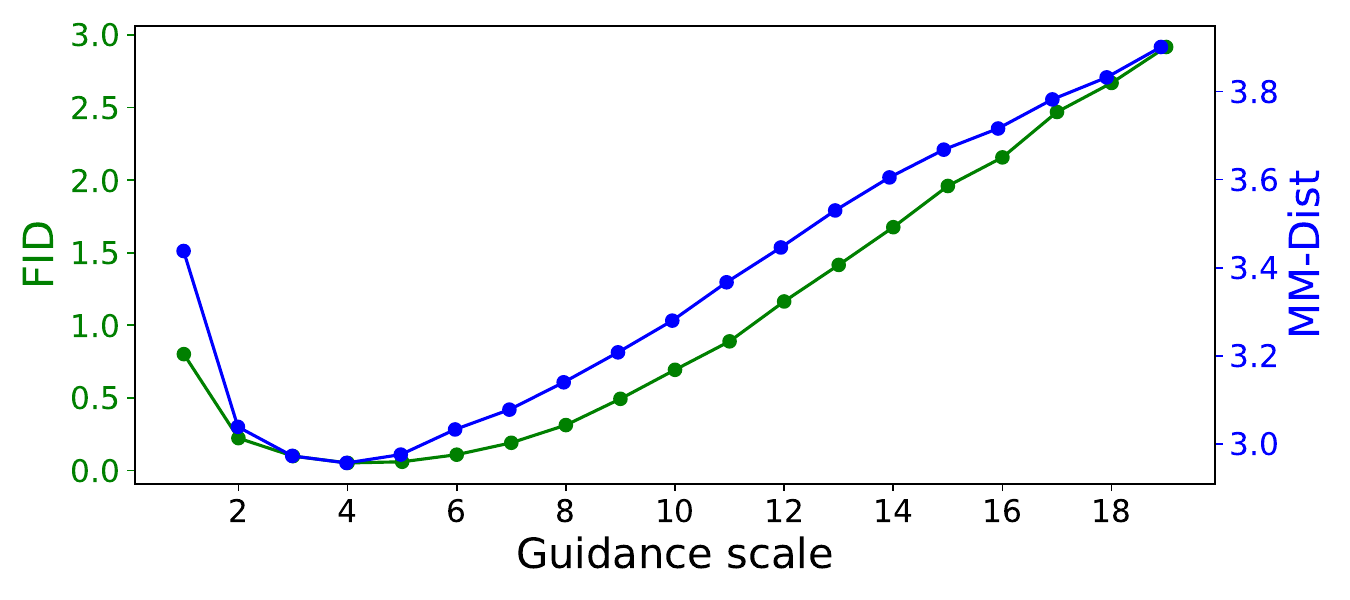}
 	\includegraphics[width=0.7\linewidth]{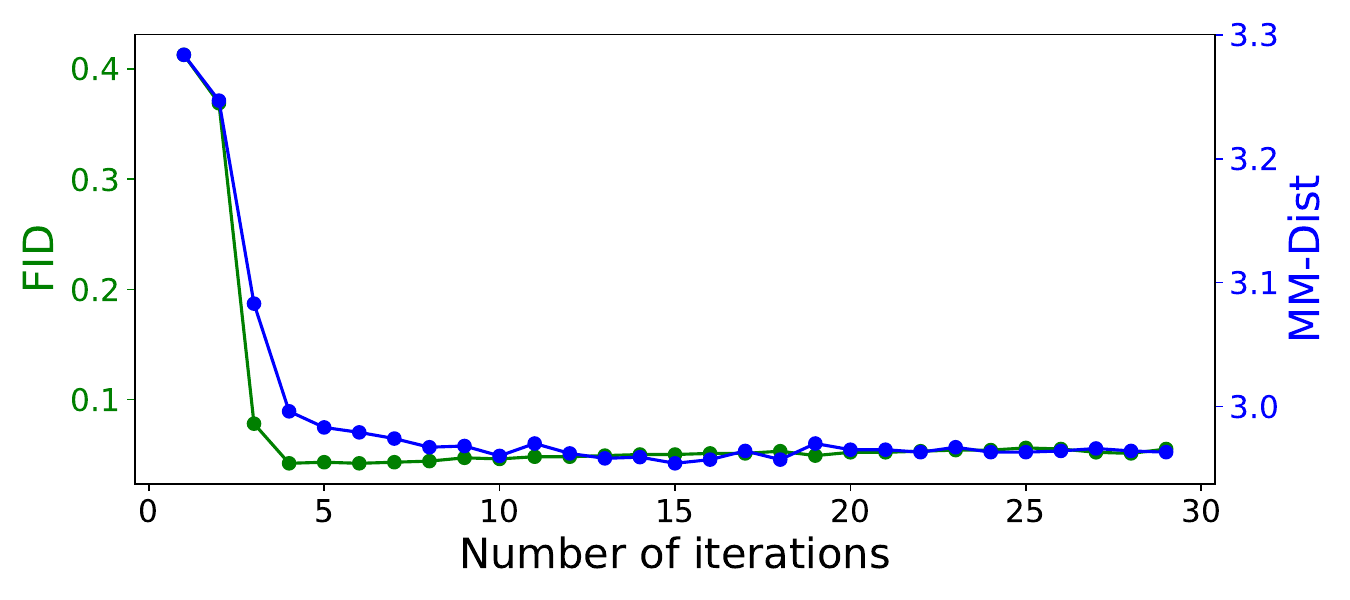}
	\caption{Evaluation sweep over guidance scale $s$ (top) and iteration numbers $L$ (bottom) in inference. We find a accuracy-fidelity sweep spot around $s=4$, meanwhile 10 iterations ($L=10$) for masked decoding yield sufficiently good results.}
        \label{fig:ablation}
\end{figure}

\subsection{Component Analysis}

In~\Cref{tab:ablation}, we comprehensively evaluate the impact of different design components in MoMask through various comparisons, showcasing the performance in both motion reconstruction and generation. Initially, we compare our approach with previous VQ-based motion generation methods~\cite{guo2022tm2t,zhang2023t2m,kong2023priority} on the HumanML3D and KIT-ML datasets. Notably, M2DM~\cite{kong2023priority} incorporates orthogonality constraints among all codebook entries to enhance VQ performance. Our residual design shows clearly superior performance when comparing with these single VQ-based approaches.

\noindent\textbf{Ablation.} In the ablation experiments, we observe that both residual quantization (RQ) and quantization dropout (QDropout) effectively contribute to the enhancement of motion quality in terms of both reconstruction and generation. Additionally,  replacing-and-remasking strategy, as well as RQ, facilitates more faithful motion generation.


\noindent\textbf{Number of Residual Layers ($V$).} In ~\cref{tab:ablation}, we investigate RVQ with different numbers of quantization layers. Generally, more residual VQ layers result in more precise reconstruction, but they also increase the burden on the R-Transformer for residual token generation. We particularly observe that the generation performance starts to degrade with more than 5 residual layers. This finding emphasizes the importance of striking a balance in the number of residual layers for optimal performance.

\noindent\textbf{Quantization Dropout ($q$).} We also analyze the impact of quantization dropout ratio $q$ in~\cref{tab:ablation}. As we increase dropout probability from 0.2, the performance gains become marginal, or even converse. We speculate that frequent disabling quantization layers may disturb the learning of quantization models.




\noindent\textbf{Inference Hyper-parameters.} The CFG scale $s$ and the number of iterations $L$ are two crucial hyperparameters during the inference of masked modeling. In~\cref{fig:ablation}, we present the performance curves of FID and multimodality distance by sweeping over different values of $s$ and $L$. Several key observations emerge. Firstly, an optimal guidance scale $s$ for M-Transformer inference is identified around $s=4$. Over-guided decoding may even inversely deteriorate the performance. Secondly, more iterations are not necessarily better. As $L$ increases, the FID and multimodality distance converge to the minima quickly, typically within around 10 iterations. Beyond 10 iterations, there are no further performance gains in both FID and multimodal distance. In this regard, our MoMask requires fewer inference steps compared to most autoregressive and diffusion models.


\subsection{Application: Temporal Inpainting}
In~\cref{fig:editing}, we showcase the capability of MoMask in temporally inpainting a specific region in a motion sequence. The region can be freely located in the middle, suffix, or prefix. Specifically, we mask out all the tokens in the region of interest and then follow the same inference procedure described in~\cref{subsec:inference}. For both tasks, our approach generates smooth motions in coherence with the given text descriptions. Additionally, we conduct a user study to quantitatively compare our inpainting results with those of MDM~\cite{tevet2022human}. In this study, 40 samples are generated from both methods using the same motion and text input, and presented to users side-by-side. With 6 users involved, 68\% of the results from MoMask are preferred over MDM.



\section{Discussion and Conclusion}
\noindent\textbf{Limitations.} We acknowledge certain limitations of MoMask. Firstly, while MoMask excels in fidelity and faithfulness for text-to-motion synthesis, its diversity is relatively limited. We plan to delve into the underlying causes of this limitation in future work. Secondly, MoMask requires the target length as input for motion generation. This could be properly addressed by applying the text2length sampling~\cite{guo2022generating} beforehand. Thirdly, akin to most VQ-based methods, MoMask may face challenges when generating motions with fast-changing root motions, such as \textit{spinning}. Exemplar cases are presented in the supplementary videos.

In conclusion, we introduce MoMask, a novel generative masked modeling framework for text-driven 3D human motion generation. MoMask features three advanced techniques: residual quantization for precise motion quantization, masked transformer and residual transformer for high-quality and faithful motion generation. MoMask is efficient and flexible, achieving superior performance without extra inference burden, and effortlessly supporting temporal motion inpainting in multiple contexts.
{
    \small
    \bibliographystyle{ieeenat_fullname}
    \bibliography{main}

\begin{thebibliography}{53}
\providecommand{\natexlab}[1]{#1}
\providecommand{\url}[1]{\texttt{#1}}
\expandafter\ifx\csname urlstyle\endcsname\relax
  \providecommand{\doi}[1]{doi: #1}\else
  \providecommand{\doi}{doi: \begingroup \urlstyle{rm}\Url}\fi

\bibitem[Ahuja and Morency(2019)]{ahuja2019language2pose}
Chaitanya Ahuja and Louis-Philippe Morency.
\newblock Language2pose: Natural language grounded pose forecasting.
\newblock In \emph{2019 International Conference on 3D Vision (3DV)}, pages 719--728. IEEE, 2019.

\bibitem[Aristidou et~al.(2018)Aristidou, Cohen-Or, Hodgins, Chrysanthou, and Shamir]{aristidou2018deep}
Andreas Aristidou, Daniel Cohen-Or, Jessica~K Hodgins, Yiorgos Chrysanthou, and Ariel Shamir.
\newblock Deep motifs and motion signatures.
\newblock \emph{ACM Transactions on Graphics (TOG)}, 37\penalty0 (6):\penalty0 1--13, 2018.

\bibitem[Athanasiou et~al.(2022)Athanasiou, Petrovich, Black, and Varol]{athanasiou2022teach}
Nikos Athanasiou, Mathis Petrovich, Michael~J Black, and G{\"u}l Varol.
\newblock Teach: Temporal action composition for 3d humans.
\newblock In \emph{2022 International Conference on 3D Vision (3DV)}, pages 414--423. IEEE, 2022.

\bibitem[Borsos et~al.(2023)Borsos, Marinier, Vincent, Kharitonov, Pietquin, Sharifi, Roblek, Teboul, Grangier, Tagliasacchi, et~al.]{borsos2023audiolm}
Zal{\'a}n Borsos, Rapha{\"e}l Marinier, Damien Vincent, Eugene Kharitonov, Olivier Pietquin, Matt Sharifi, Dominik Roblek, Olivier Teboul, David Grangier, Marco Tagliasacchi, et~al.
\newblock Audiolm: a language modeling approach to audio generation.
\newblock \emph{IEEE/ACM Transactions on Audio, Speech, and Language Processing}, 2023.

\bibitem[Cai et~al.(2018)Cai, Bai, Tai, and Tang]{cai2018deep}
Haoye Cai, Chunyan Bai, Yu-Wing Tai, and Chi-Keung Tang.
\newblock Deep video generation, prediction and completion of human action sequences.
\newblock In \emph{Proceedings of the European Conference on Computer Vision (ECCV)}, pages 366--382, 2018.

\bibitem[Cervantes et~al.(2022)Cervantes, Sekikawa, Sato, and Shinoda]{cervantes2022implicit}
Pablo Cervantes, Yusuke Sekikawa, Ikuro Sato, and Koichi Shinoda.
\newblock Implicit neural representations for variable length human motion generation.
\newblock In \emph{European Conference on Computer Vision}, pages 356--372. Springer, 2022.

\bibitem[Chang et~al.(2022)Chang, Zhang, Jiang, Liu, and Freeman]{chang2022maskgit}
Huiwen Chang, Han Zhang, Lu Jiang, Ce Liu, and William~T Freeman.
\newblock Maskgit: Masked generative image transformer.
\newblock In \emph{Proceedings of the IEEE/CVF Conference on Computer Vision and Pattern Recognition}, pages 11315--11325, 2022.

\bibitem[Chang et~al.(2023)Chang, Zhang, Barber, Maschinot, Lezama, Jiang, Yang, Murphy, Freeman, Rubinstein, et~al.]{chang2023muse}
Huiwen Chang, Han Zhang, Jarred Barber, AJ Maschinot, Jose Lezama, Lu Jiang, Ming-Hsuan Yang, Kevin Murphy, William~T Freeman, Michael Rubinstein, et~al.
\newblock Muse: Text-to-image generation via masked generative transformers.
\newblock \emph{arXiv preprint arXiv:2301.00704}, 2023.

\bibitem[Chen et~al.(2023)Chen, Jiang, Liu, Huang, Fu, Chen, and Yu]{chen2023executing}
Xin Chen, Biao Jiang, Wen Liu, Zilong Huang, Bin Fu, Tao Chen, and Gang Yu.
\newblock Executing your commands via motion diffusion in latent space.
\newblock In \emph{Proceedings of the IEEE/CVF Conference on Computer Vision and Pattern Recognition}, pages 18000--18010, 2023.

\bibitem[Devlin et~al.(2018)Devlin, Chang, Lee, and Toutanova]{devlin2018bert}
Jacob Devlin, Ming-Wei Chang, Kenton Lee, and Kristina Toutanova.
\newblock Bert: Pre-training of deep bidirectional transformers for language understanding.
\newblock \emph{arXiv preprint arXiv:1810.04805}, 2018.

\bibitem[Ferdowsi et~al.(2017)Ferdowsi, Voloshynovskiy, and Kostadinov]{ferdowsi2017regularized}
Sohrab Ferdowsi, Slava Voloshynovskiy, and Dimche Kostadinov.
\newblock Regularized residual quantization: a multi-layer sparse dictionary learning approach.
\newblock \emph{arXiv preprint arXiv:1705.00522}, 2017.

\bibitem[Ghosh et~al.(2021)Ghosh, Cheema, Oguz, Theobalt, and Slusallek]{ghosh2021synthesis}
Anindita Ghosh, Noshaba Cheema, Cennet Oguz, Christian Theobalt, and Philipp Slusallek.
\newblock Synthesis of compositional animations from textual descriptions.
\newblock In \emph{Proceedings of the IEEE/CVF International Conference on Computer Vision}, pages 1396--1406, 2021.

\bibitem[Gong et~al.(2023)Gong, Lian, Chang, Guo, Zuo, Jiang, and Wang]{gong2023tm2d}
Kehong Gong, Dongze Lian, Heng Chang, Chuan Guo, Xinxin Zuo, Zihang Jiang, and Xinchao Wang.
\newblock Tm2d: Bimodality driven 3d dance generation via music-text integration.
\newblock \emph{arXiv preprint arXiv:2304.02419}, 2023.

\bibitem[Guo et~al.(2020)Guo, Zuo, Wang, Zou, Sun, Deng, Gong, and Cheng]{guo2020action2motion}
Chuan Guo, Xinxin Zuo, Sen Wang, Shihao Zou, Qingyao Sun, Annan Deng, Minglun Gong, and Li Cheng.
\newblock Action2motion: Conditioned generation of 3d human motions.
\newblock In \emph{Proceedings of the 28th ACM International Conference on Multimedia}, pages 2021--2029, 2020.

\bibitem[Guo et~al.(2022{\natexlab{a}})Guo, Zou, Zuo, Wang, Ji, Li, and Cheng]{guo2022generating}
Chuan Guo, Shihao Zou, Xinxin Zuo, Sen Wang, Wei Ji, Xingyu Li, and Li Cheng.
\newblock Generating diverse and natural 3d human motions from text.
\newblock In \emph{Proceedings of the IEEE/CVF Conference on Computer Vision and Pattern Recognition}, pages 5152--5161, 2022{\natexlab{a}}.

\bibitem[Guo et~al.(2022{\natexlab{b}})Guo, Zuo, Wang, and Cheng]{guo2022tm2t}
Chuan Guo, Xinxin Zuo, Sen Wang, and Li Cheng.
\newblock Tm2t: Stochastic and tokenized modeling for the reciprocal generation of 3d human motions and texts.
\newblock In \emph{European Conference on Computer Vision}, pages 580--597. Springer, 2022{\natexlab{b}}.

\bibitem[Guo et~al.(2022{\natexlab{c}})Guo, Zuo, Wang, Liu, Zou, Gong, and Cheng]{guo2022action2video}
Chuan Guo, Xinxin Zuo, Sen Wang, Xinshuang Liu, Shihao Zou, Minglun Gong, and Li Cheng.
\newblock Action2video: Generating videos of human 3d actions.
\newblock \emph{International Journal of Computer Vision}, 130\penalty0 (2):\penalty0 285--315, 2022{\natexlab{c}}.

\bibitem[He et~al.(2022)He, Chen, Xie, Li, Doll{\'a}r, and Girshick]{he2022masked}
Kaiming He, Xinlei Chen, Saining Xie, Yanghao Li, Piotr Doll{\'a}r, and Ross Girshick.
\newblock Masked autoencoders are scalable vision learners.
\newblock In \emph{Proceedings of the IEEE/CVF Conference on Computer Vision and Pattern Recognition}, pages 16000--16009, 2022.

\bibitem[Ho and Salimans(2022)]{ho2022classifier}
Jonathan Ho and Tim Salimans.
\newblock Classifier-free diffusion guidance.
\newblock \emph{arXiv preprint arXiv:2207.12598}, 2022.

\bibitem[Huang et~al.(2020)Huang, Hu, Wu, Sawada, Zhang, and Jiang]{huang2020dance}
Ruozi Huang, Huang Hu, Wei Wu, Kei Sawada, Mi Zhang, and Daxin Jiang.
\newblock Dance revolution: Long-term dance generation with music via curriculum learning.
\newblock \emph{arXiv preprint arXiv:2006.06119}, 2020.

\bibitem[Jiang et~al.(2023)Jiang, Chen, Liu, Yu, Yu, and Chen]{jiang2023motiongpt}
Biao Jiang, Xin Chen, Wen Liu, Jingyi Yu, Gang Yu, and Tao Chen.
\newblock Motiongpt: Human motion as a foreign language.
\newblock \emph{arXiv preprint arXiv:2306.14795}, 2023.

\bibitem[Kim et~al.(2023)Kim, Kim, and Choi]{kim2023flame}
Jihoon Kim, Jiseob Kim, and Sungjoon Choi.
\newblock Flame: Free-form language-based motion synthesis \& editing.
\newblock In \emph{Proceedings of the AAAI Conference on Artificial Intelligence}, pages 8255--8263, 2023.

\bibitem[Kong et~al.(2023)Kong, Gong, Lian, Mi, and Wang]{kong2023priority}
Hanyang Kong, Kehong Gong, Dongze Lian, Michael~Bi Mi, and Xinchao Wang.
\newblock Priority-centric human motion generation in discrete latent space.
\newblock In \emph{Proceedings of the IEEE/CVF International Conference on Computer Vision}, pages 14806--14816, 2023.

\bibitem[Li et~al.(2023)Li, Chang, Mishra, Zhang, Katabi, and Krishnan]{li2023mage}
Tianhong Li, Huiwen Chang, Shlok Mishra, Han Zhang, Dina Katabi, and Dilip Krishnan.
\newblock Mage: Masked generative encoder to unify representation learning and image synthesis.
\newblock In \emph{Proceedings of the IEEE/CVF Conference on Computer Vision and Pattern Recognition}, pages 2142--2152, 2023.

\bibitem[Li et~al.(2021)Li, Ding, Liu, Zhang, and Guo]{li2021trq}
Yue Li, Wenrui Ding, Chunlei Liu, Baochang Zhang, and Guodong Guo.
\newblock Trq: Ternary neural networks with residual quantization.
\newblock In \emph{Proceedings of the AAAI Conference on Artificial Intelligence}, pages 8538--8546, 2021.

\bibitem[Li et~al.(2017)Li, Ni, Zhang, Yang, and Gao]{li2017performance}
Zefan Li, Bingbing Ni, Wenjun Zhang, Xiaokang Yang, and Wen Gao.
\newblock Performance guaranteed network acceleration via high-order residual quantization.
\newblock In \emph{Proceedings of the IEEE International Conference on Computer Vision}, pages 2584--2592, 2017.

\bibitem[Lin et~al.(2018)Lin, Wu, Corona, Tai, Huang, and Mooney]{lin2018generating}
Angela~S Lin, Lemeng Wu, Rodolfo Corona, Kevin Tai, Qixing Huang, and Raymond~J Mooney.
\newblock Generating animated videos of human activities from natural language descriptions.
\newblock \emph{Learning}, 2018\penalty0 (1), 2018.

\bibitem[Lin et~al.(2023)Lin, Chang, Liu, Li, Lin, Tian, and Chen]{lin2023being}
Junfan Lin, Jianlong Chang, Lingbo Liu, Guanbin Li, Liang Lin, Qi Tian, and Chang-wen Chen.
\newblock Being comes from not-being: Open-vocabulary text-to-motion generation with wordless training.
\newblock In \emph{Proceedings of the IEEE/CVF Conference on Computer Vision and Pattern Recognition}, pages 23222--23231, 2023.

\bibitem[Liu et~al.(2022)Liu, Wu, Jin, Ji, Liu, Lu, and Cheng]{liu2022investigating}
Zhenguang Liu, Shuang Wu, Shuyuan Jin, Shouling Ji, Qi Liu, Shijian Lu, and Li Cheng.
\newblock Investigating pose representations and motion contexts modeling for 3d motion prediction.
\newblock \emph{IEEE Transactions on Pattern Analysis and Machine Intelligence}, 45\penalty0 (1):\penalty0 681--697, 2022.

\bibitem[Lou et~al.(2023)Lou, Zhu, Wang, Wang, and Yang]{lou2023diversemotion}
Yunhong Lou, Linchao Zhu, Yaxiong Wang, Xiaohan Wang, and Yi Yang.
\newblock Diversemotion: Towards diverse human motion generation via discrete diffusion.
\newblock \emph{arXiv preprint arXiv:2309.01372}, 2023.

\bibitem[Lucas et~al.(2022)Lucas, Baradel, Weinzaepfel, and Rogez]{lucas2022posegpt}
Thomas Lucas, Fabien Baradel, Philippe Weinzaepfel, and Gr{\'e}gory Rogez.
\newblock Posegpt: Quantization-based 3d human motion generation and forecasting.
\newblock In \emph{European Conference on Computer Vision}, pages 417--435. Springer, 2022.

\bibitem[Mahmood et~al.(2019)Mahmood, Ghorbani, Troje, Pons-Moll, and Black]{mahmood2019amass}
Naureen Mahmood, Nima Ghorbani, Nikolaus~F Troje, Gerard Pons-Moll, and Michael~J Black.
\newblock Amass: Archive of motion capture as surface shapes.
\newblock In \emph{Proceedings of the IEEE/CVF International Conference on Computer Vision}, pages 5442--5451, 2019.

\bibitem[Mao et~al.(2019)Mao, Liu, Salzmann, and Li]{mao2019learning}
Wei Mao, Miaomiao Liu, Mathieu Salzmann, and Hongdong Li.
\newblock Learning trajectory dependencies for human motion prediction.
\newblock In \emph{Proceedings of the IEEE/CVF International Conference on Computer Vision}, pages 9489--9497, 2019.

\bibitem[Martinez et~al.(2014)Martinez, Hoos, and Little]{martinez2014stacked}
Julieta Martinez, Holger~H Hoos, and James~J Little.
\newblock Stacked quantizers for compositional vector compression.
\newblock \emph{arXiv preprint arXiv:1411.2173}, 2014.

\bibitem[Petrovich et~al.(2021)Petrovich, Black, and Varol]{petrovich2021action}
Mathis Petrovich, Michael~J Black, and G{\"u}l Varol.
\newblock Action-conditioned 3d human motion synthesis with transformer vae.
\newblock In \emph{Proceedings of the IEEE/CVF International Conference on Computer Vision}, pages 10985--10995, 2021.

\bibitem[Petrovich et~al.(2022)Petrovich, Black, and Varol]{petrovich2022temos}
Mathis Petrovich, Michael~J Black, and G{\"u}l Varol.
\newblock Temos: Generating diverse human motions from textual descriptions.
\newblock In \emph{European Conference on Computer Vision}, pages 480--497. Springer, 2022.

\bibitem[Plappert et~al.(2016)Plappert, Mandery, and Asfour]{plappert2016kit}
Matthias Plappert, Christian Mandery, and Tamim Asfour.
\newblock The kit motion-language dataset.
\newblock \emph{Big data}, 4\penalty0 (4):\penalty0 236--252, 2016.

\bibitem[Plappert et~al.(2018)Plappert, Mandery, and Asfour]{plappert2018learning}
Matthias Plappert, Christian Mandery, and Tamim Asfour.
\newblock Learning a bidirectional mapping between human whole-body motion and natural language using deep recurrent neural networks.
\newblock \emph{Robotics and Autonomous Systems}, 109:\penalty0 13--26, 2018.

\bibitem[Radford et~al.(2021)Radford, Kim, Hallacy, Ramesh, Goh, Agarwal, Sastry, Askell, Mishkin, Clark, et~al.]{radford2021learning}
Alec Radford, Jong~Wook Kim, Chris Hallacy, Aditya Ramesh, Gabriel Goh, Sandhini Agarwal, Girish Sastry, Amanda Askell, Pamela Mishkin, Jack Clark, et~al.
\newblock Learning transferable visual models from natural language supervision.
\newblock In \emph{International Conference on Machine Learning}, pages 8748--8763. PMLR, 2021.

\bibitem[Siyao et~al.(2022)Siyao, Yu, Gu, Lin, Wang, Qian, Loy, and Liu]{siyao2022bailando}
Li Siyao, Weijiang Yu, Tianpei Gu, Chunze Lin, Quan Wang, Chen Qian, Chen~Change Loy, and Ziwei Liu.
\newblock Bailando: 3d dance generation by actor-critic gpt with choreographic memory.
\newblock In \emph{Proceedings of the IEEE/CVF Conference on Computer Vision and Pattern Recognition}, pages 11050--11059, 2022.

\bibitem[Tevet et~al.(2022{\natexlab{a}})Tevet, Gordon, Hertz, Bermano, and Cohen-Or]{tevet2022motionclip}
Guy Tevet, Brian Gordon, Amir Hertz, Amit~H Bermano, and Daniel Cohen-Or.
\newblock Motionclip: Exposing human motion generation to clip space.
\newblock In \emph{European Conference on Computer Vision}, pages 358--374. Springer, 2022{\natexlab{a}}.

\bibitem[Tevet et~al.(2022{\natexlab{b}})Tevet, Raab, Gordon, Shafir, Cohen-Or, and Bermano]{tevet2022human}
Guy Tevet, Sigal Raab, Brian Gordon, Yonatan Shafir, Daniel Cohen-Or, and Amit~H Bermano.
\newblock Human motion diffusion model.
\newblock \emph{arXiv preprint arXiv:2209.14916}, 2022{\natexlab{b}}.

\bibitem[Tseng et~al.(2023)Tseng, Castellon, and Liu]{tseng2023edge}
Jonathan Tseng, Rodrigo Castellon, and Karen Liu.
\newblock Edge: Editable dance generation from music.
\newblock In \emph{Proceedings of the IEEE/CVF Conference on Computer Vision and Pattern Recognition}, pages 448--458, 2023.

\bibitem[Van Den~Oord et~al.(2017)Van Den~Oord, Vinyals, et~al.]{van2017neural}
Aaron Van Den~Oord, Oriol Vinyals, et~al.
\newblock Neural discrete representation learning.
\newblock \emph{Advances in Neural Information Processing Systems}, 30, 2017.

\bibitem[Wang et~al.(2023)Wang, Chen, Wu, Zhang, Zhou, Liu, Chen, Liu, Wang, Li, et~al.]{wang2023neural}
Chengyi Wang, Sanyuan Chen, Yu Wu, Ziqiang Zhang, Long Zhou, Shujie Liu, Zhuo Chen, Yanqing Liu, Huaming Wang, Jinyu Li, et~al.
\newblock Neural codec language models are zero-shot text to speech synthesizers.
\newblock \emph{arXiv preprint arXiv:2301.02111}, 2023.

\bibitem[Wang et~al.(2020)Wang, Yu, Zhao, Zhang, Zhou, Yuan, and Chen]{wang2020learning}
Zhenyi Wang, Ping Yu, Yang Zhao, Ruiyi Zhang, Yufan Zhou, Junsong Yuan, and Changyou Chen.
\newblock Learning diverse stochastic human-action generators by learning smooth latent transitions.
\newblock In \emph{Proceedings of the AAAI Conference on Artificial Intelligence}, pages 12281--12288, 2020.

\bibitem[Yu et~al.(2023)Yu, Cheng, Sohn, Lezama, Zhang, Chang, Hauptmann, Yang, Hao, Essa, et~al.]{yu2023magvit}
Lijun Yu, Yong Cheng, Kihyuk Sohn, Jos{\'e} Lezama, Han Zhang, Huiwen Chang, Alexander~G Hauptmann, Ming-Hsuan Yang, Yuan Hao, Irfan Essa, et~al.
\newblock Magvit: Masked generative video transformer.
\newblock In \emph{Proceedings of the IEEE/CVF Conference on Computer Vision and Pattern Recognition}, pages 10459--10469, 2023.

\bibitem[Zeghidour et~al.(2021)Zeghidour, Luebs, Omran, Skoglund, and Tagliasacchi]{zeghidour2021soundstream}
Neil Zeghidour, Alejandro Luebs, Ahmed Omran, Jan Skoglund, and Marco Tagliasacchi.
\newblock Soundstream: An end-to-end neural audio codec.
\newblock \emph{IEEE/ACM Transactions on Audio, Speech, and Language Processing}, 30:\penalty0 495--507, 2021.

\bibitem[Zhang et~al.(2023{\natexlab{a}})Zhang, Zhang, Cun, Huang, Zhang, Zhao, Lu, and Shen]{zhang2023t2m}
Jianrong Zhang, Yangsong Zhang, Xiaodong Cun, Shaoli Huang, Yong Zhang, Hongwei Zhao, Hongtao Lu, and Xi Shen.
\newblock T2m-gpt: Generating human motion from textual descriptions with discrete representations.
\newblock \emph{arXiv preprint arXiv:2301.06052}, 2023{\natexlab{a}}.

\bibitem[Zhang et~al.(2022)Zhang, Cai, Pan, Hong, Guo, Yang, and Liu]{zhang2022motiondiffuse}
Mingyuan Zhang, Zhongang Cai, Liang Pan, Fangzhou Hong, Xinying Guo, Lei Yang, and Ziwei Liu.
\newblock Motiondiffuse: Text-driven human motion generation with diffusion model.
\newblock \emph{arXiv preprint arXiv:2208.15001}, 2022.

\bibitem[Zhang et~al.(2023{\natexlab{b}})Zhang, Guo, Pan, Cai, Hong, Li, Yang, and Liu]{zhang2023remodiffuse}
Mingyuan Zhang, Xinying Guo, Liang Pan, Zhongang Cai, Fangzhou Hong, Huirong Li, Lei Yang, and Ziwei Liu.
\newblock Remodiffuse: Retrieval-augmented motion diffusion model.
\newblock \emph{arXiv preprint arXiv:2304.01116}, 2023{\natexlab{b}}.

\bibitem[Zhang et~al.(2023{\natexlab{c}})Zhang, Huang, Liu, Tang, Lu, Chen, Bai, Chu, Yu, and Ouyang]{zhang2023motiongpt}
Yaqi Zhang, Di Huang, Bin Liu, Shixiang Tang, Yan Lu, Lu Chen, Lei Bai, Qi Chu, Nenghai Yu, and Wanli Ouyang.
\newblock Motiongpt: Finetuned llms are general-purpose motion generators.
\newblock \emph{arXiv preprint arXiv:2306.10900}, 2023{\natexlab{c}}.

\bibitem[Zhou and Wang(2023)]{zhou2023ude}
Zixiang Zhou and Baoyuan Wang.
\newblock Ude: A unified driving engine for human motion generation.
\newblock In \emph{Proceedings of the IEEE/CVF Conference on Computer Vision and Pattern Recognition}, pages 5632--5641, 2023.

\end{thebibliography}
}


\end{document}